%% file: paper.tex
\documentclass[nohyperref]{article}

\usepackage{microtype}
\usepackage{nicefrac}       
\usepackage{graphicx}
\usepackage{booktabs} 
\usepackage{caption}
\usepackage{subcaption}
\usepackage{xcolor}
\usepackage{amsmath,bm,multicol,multirow}
\usepackage{pgfplotstable}
\usepackage[scientific-notation=true]{siunitx}

\usepgfplotslibrary{fillbetween}
\usetikzlibrary[patterns]

\usepackage{hyperref}

\usepackage{ulem}
\usepackage[accepted]{icml2022}

\usepackage{amssymb}
\usepackage{mathtools}
\usepackage{amsthm}

\usepackage[capitalize,noabbrev]{cleveref}

\theoremstyle{plain}

\theoremstyle{definition}

\theoremstyle{remark}

\usepackage[textsize=tiny]{todonotes}

\icmltitlerunning{\name{}: A General Framework for Self-supervised Learning in Speech, Vision and Language}

\input{macro}
\input{tables}

\begin{document}

\twocolumn[
\icmltitle{\name{}: A General Framework for Self-supervised Learning in Speech, Vision and Language}



\icmlsetsymbol{equal}{*}

\begin{icmlauthorlist}
\icmlauthor{Alexei Baevski}{meta}
\icmlauthor{Wei-Ning Hsu}{meta}
\icmlauthor{Qiantong Xu}{samba}
\icmlauthor{Arun Babu}{meta}
\icmlauthor{Jiatao Gu}{meta}
\icmlauthor{Michael Auli}{meta}
\end{icmlauthorlist}

\icmlaffiliation{meta}{Meta AI}
\icmlaffiliation{samba}{SambaNova, work done while at Meta AI}

\icmlcorrespondingauthor{Alexei Baevski}{abaevski@fb.com}
\icmlcorrespondingauthor{Michael Auli}{michaelauli@fb.com}

\icmlkeywords{Machine Learning, ICML}

\vskip 0.3in
]



\printAffiliationsAndNotice{}  

\begin{abstract}
While the general idea of self-supervised learning is identical across modalities, the actual algorithms and objectives differ widely because they were developed with a single modality in mind.
To get us closer to general self-supervised learning, we present \name{}, a framework that uses the same learning method for either speech, NLP or computer vision.
The core idea is to predict latent representations of the full input data based on a masked view of the input in a self-distillation setup using a standard Transformer architecture.  
Instead of predicting modality-specific targets such as words, visual tokens or units of human speech which are local in nature, \name{} predicts contextualized latent representations that contain information from the entire input.
Experiments on the major benchmarks of speech recognition, image classification, and natural language understanding demonstrate a new state of the art or competitive performance to predominant approaches.
Models and code are available at {\small\url{www.github.com/pytorch/fairseq/tree/master/examples/data2vec}}.
\end{abstract}

\section{Introduction}
\label{sec:intro}

Self-supervised learning builds representations of data without human annotated labels which led to significant advances in natural language processing (NLP;~\citealt{peters2018acl,radford2018unsup,devlin2018bert,brown2020gpt3}), speech processing~\citep{oord2018cpc,schneider2019wav2vec,baevski2020wav} as well as computer vision~\citep{chen2020simple,chen2021mocov3,caron2021dino,bao2021beit,he2021mae}.
Self-supervised representations have even enabled completely unsupervised learning in tasks such as machine translation~\citep{lample2018unsupmt} and speech recognition~\citep{baevski2021unsupervised}.

Research in self-supervised algorithms has focused on individual modalities which results in specific designs and learning biases.
For example, in speech processing, there is no vocabulary of speech units over which we can define a self-supervised learning task such as words in NLP\footnote{This is true for many languages but not for certain Asian languages.} and therefore several prominent models are equipped with mechanisms to learn an inventory of speech units~\citep{baevski2020wav,hsu2020hubert}.
A similar problem exists for computer vision, where researchers either learn discrete visual tokens~\citep{radford2021clip,bao2021beit}, regress the input~\citep{he2021mae} or learn representations invariant to data augmentation~\citep{chen2020simple,grill2020byol,caron2021dino}.

\begin{figure*}[h!t]
\begin{center}
\includegraphics[width=\linewidth]{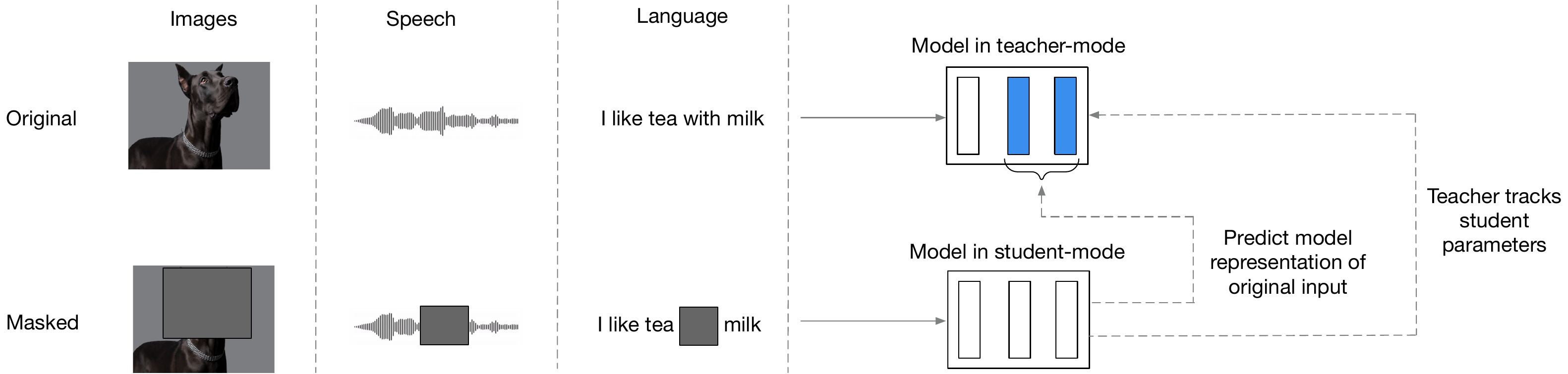}
\caption{Illustration of how \name{} follows the same learning process for different modalities. 
The model first produces representations of the original input example (teacher mode) which are then regressed by the same model based on a masked version of the input. 
The teacher parameters are an exponentially moving average of the student weights. 
The student predicts the average of $K$ network layers of the teacher (shaded in blue). 
\label{fig:model}
}
\end{center}
\end{figure*}

While learning biases are certainly helpful, it is often unclear whether they will generalize to other modalities.
Moreover, leading theories on the biology of learning~\citep{friston2009predictive,friston2010free} imply that humans likely use similar learning processes to understand the visual world as they do for language.
Relatedly, general neural network architectures have been shown to perform very well compared to modality-specific counterparts~\citep{jaegle2021perceiver}.

In an effort to get closer to machines that learn in general ways about the environment, we designed \name{}, a framework for general self-supervised learning that works for images, speech and text where the learning objective is identical in each modality.
The present work unifies the learning algorithm but still learns representations individually for each modality.
We hope that a single algorithm will make future multi-modal learning simpler, more effective and lead to models that understand the world better through multiple modalities.

Our method combines masked prediction~\citep{devlin2018bert,baevski2020wav,bao2021beit} with the learning of latent target representations~\citep{grill2020byol,caron2021dino} but generalizes the latter by using multiple network layers as targets and shows that this approach works across several modalities.
Specifically, we train an off-the-shelf Transformer network~\citep{vaswani2017transformer} which we use either in teacher or student mode (Illustration in~\autoref{fig:model}):
we first build representations of the full input data whose purpose is to serve as targets in the learning task (teacher mode).
Next, we encode a masked version of the input sample with which we predict the full data representations (student mode).
The weights of the teacher are an exponentially decaying average of the student~\citep{he2019momentum,grill2020byol,caron2021dino}.
Since different modalities have vastly different inputs, e.g., pixels vs. words, we use modality-specific feature encoders and masking strategies from the literature.

Since our method works with the latent network representations of the learner itself, it can be seen as a simplification of many modality-specific designs such as learning a fixed set of visual tokens~\citep{radford2021clip,oord2017vqvae}, or normalization of the input to create suitable targets~\citep{he2021mae}, or the learning of a vocabulary of discrete speech units~\citep{baevski2020wav,hsu2020hubert}. 
Moreover, our target representations are \emph{continuous} and \emph{contextualized}, through the use of self-attention, which makes them richer than a fixed set of targets and/or targets based on local context such as used in most prior work.

Experimental results show \name{} to be effective in all three modalities, setting a new state of the art for ViT-B with single models and ViT-L on ImageNet-1K, improving over the best prior work in speech processing on speech recognition~\citep{baevski2020wav,hsu2020hubert} and outperforming a like for like RoBERTa baseline on the GLUE natural language understanding benchmark~\citep{liu2019roberta}.

\section{Related Work}
\label{sec:related}

\paragraph{Self-supervised Learning in Computer Vision.}
Unsupervised pre-training for computer vision has been a very active area of research with methods contrasting representations of augmentations of the same image, entirely different images~\citep{chen2020simple,grill2020byol,caron2021dino,chen2021mocov3} as well as online clustering~\citep{caron2020swav}.
Similar to our work, both BYOL~\citep{grill2020byol} and DINO~\citep{caron2021dino} regress neural network representations of a momentum encoder, but our work differs in that it uses a masked prediction task and we regress multiple neural network layer representations instead of just the top layer which we find to be more effective.
Moreover, \name{} works for multiple modalities.

The most recent work focuses on training vision Transformers~\citep{dosovitskiy2020vit} with masked prediction objectives~\citep{bao2021beit,he2021mae,xie2021simmim} whose performance surpasses supervised-only training on ImageNet-1K.
Several of these methods predict visual tokens~\citep{bao2021beit,he2021mae,dong2022peco} learned in a separate step before pre-training~\citep{oord2017vqvae,ramesh2021zero}, during pretraining~\citep{zhou2021ibot}, and others directly predict the input pixels~\citep{he2021mae,xie2021simmim}.

Instead, \name{} predicts the latent representations of the input data.
Another difference to this body of work is that the latent target representations are \emph{contextualized}, incorporating relevant features from the entire image instead of targets which contain information isolated to the current patch, such as visual tokens or pixels.

\paragraph{Self-supervised Learning in NLP.}
Pre-training has been very successful in advancing natural language understanding~\citep{mccann2017cove,peters2018acl,radford2018unsup,baevski2019bitransformer,devlin2018bert,yang2019xlnet,brown2020gpt3}. 
The most prominent model is BERT~\citep{devlin2018bert} which solves a masked prediction task where some of the input tokens are blanked out in order to be predicted given the remaining input.
For many languages it is easy to determine word boundaries and most methods therefore predict word or sub-word units for pre-training.
There is also work on knowledge distillation to obtain smaller BERT-style models, both for pre-training and fine-tuning~\citep{jiao2020tinybert}.

Compared to prior NLP algorithms, \name{} does not predict discrete linguistic tokens such as words, sub-words or bytes but rather a continuous and contextualized representation. 
This has two advantages: 
first, the targets themselves are not predefined, nor is their number limited.
This enables the model to adapt to a particular input example.
Second, targets are \emph{contextualized}, taking context information into account. 
This is unlike BERT-style models which learn a single embedding for each target which needs to fit all instances of a particular target in the data.

\paragraph{Self-supervised Learning in Speech.}
Work in self-supervised learning for speech includes autoregressive models~\citep{oord2018cpc,schneider2019wav2vec,baevski2019vqwav2vec,chung2019apc} as well as bi-directional models~\citep{baevski2020wav,hsu2020hubert,ao2021speecht5,chen2021wavlm}.
Two prominent models, wav2vec 2.0 and HuBERT are based on predicting discrete units of speech, either learned jointly during pre-training~\citep{baevski2020wav}, or in an iterative pipeline approach~\citep{hsu2020hubert} where pre-training and clustering alternate.\footnote{Quantization is optional for wav2vec 2.0~\citep{baevski2020wav,zhang2020pushing} but helpful for noisy speech~\citep{chung2021w2vbert}.
}
Another line of work directly reconstructs the input features~\citep{eloff2019unsupervised,liu2021tera}.

In comparison to wav2vec 2.0, \name{} directly predicts contextualized latent representations without quantization. 
HuBERT discretizes representations from different layers across iterations and predicts these discretized units whereas \name{} predicts the average over multiple layers.
Similar to other modalities, there is work on distilling larger self-supervised models into smaller models but primarily for the purpose of efficiency~\citep{chang2021distilhubert}.

\paragraph{Multimodal Pre-training.}
There has been a considerable body of research on learning representations of multiple modalities simultaneously often using paired data~\citep{aytar2017see,radford2021learning,wang2021vlmo,singh2021flava} with the aim to produce cross-modal representations which can perform well on multi-modal tasks and with modalities benefiting from each other through joint training~\citep{alayrac2020selfsupervised,akbari2021vatt} with recent methods exploring few-shot learning~\citep{tsimpoukelli2021multimodal}.
Our work does not perform multimodal training but aims to unifiy the learning objective for self-supervised learning in different modalities. 
We hope that this will enable better multimodal representations in the future.

\section{Method}

\name{} is trained by predicting the model representations of the full input data given a partial view of the input (\autoref{fig:model}).
We first encode a masked version of the training sample (model in \emph{student mode}) and then construct training targets by encoding the unmasked version of the input with the same model but when parameterized as an exponentially moving average of the model weights (model in \emph{teacher mode};~\citealt{grill2020byol,caron2021dino}).
The target representations encode all of the information in the training sample and the learning task is for the student to predict these representations given a partial view of the input.

\subsection{Model Architecture}

We use the standard Transformer architecture~\citep{vaswani2017transformer} with a modality-specific encoding of the input data borrowed from prior work:\footnote{While we used Transformer networks, alternative architectures may be equally applicable.}
for computer vision, we use the ViT-strategy of encoding an image as a sequence of patches, each spanning 16x16 pixels, input to a linear transformation~\citep{dosovitskiy2020vit,bao2021beit}.
Speech data is encoded using a multi-layer 1-D convolutional neural network that maps 16 kHz waveform to 50 Hz representations~\citep{baevski2020wav}.
Text is pre-processed to obtain sub-word units~\citep{sennrich2016bpe,devlin2018bert}, which are then embedded in distributional space via learned embedding vectors.
We detail these methods below (\textsection\ref{sec:setup}).

\subsection{Masking}

After the input sample has been embedded as a sequence of tokens, we mask part of these units by replacing them with a learned MASK embedding token and feed the sequence to the Transformer network.
For computer vision, we follow the block-wise masking strategy of~\citet{bao2021beit}, for speech we mask spans of latent speech representations~\citep{baevski2020wav} and for language we mask tokens~\citep{devlin2018bert}; \textsection\ref{sec:setup} details each strategy.

\subsection{Training Targets}
\label{sec:method_targets}

The model is trained to predict the model representations of the original unmasked training sample based on an encoding of the masked sample.
We predict model representations only for time-steps which are masked.
The representations we predict are \emph{contextualized representations}, encoding the particular time-step but also other information from the sample due to the use of self-attention in the Transformer network.\footnote{
In preliminary experiments, we found that additional context information for the targets was helpful since masking some of the time-steps when in teacher mode resulted in lower accuracy.
}
This is an important difference to BERT~\citep{devlin2018bert}, wav2vec 2.0~\citep{baevski2020wav} or BEiT, MAE, SimMIM, and MaskFeat~\citep{bao2021beit,he2021mae,xie2021simmim,wei2021masked} which predict targets lacking contextual information.
Below, we detail how we parameterize the teacher which predicts the network representations that will serve as targets as well as how we construct the final target vectors to be predicted by the model in student-mode.

\paragraph{Teacher Parameterization.}
The encoding of the unmasked training sample is parameterized by an exponentially moving average (EMA) of the model parameters ($\theta$;~\citealt{tarvainen2018mean,grill2020byol,caron2021dino}) where the weights of the model in target-mode $\Delta$ are:
$$
\Delta \leftarrow \tau \Delta + (1 - \tau)~\theta
$$

We use a schedule for $\tau$ that linearly increases this parameter  from $\tau_0$ to the target value $\tau_e$ over the first $\tau_n$ updates after which the value is kept constant for the remainder of training.
This strategy results in the teacher being updated more frequently at the beginning of training, when the model is random, and less frequently later in training, when good parameters have already been learned. 
We found it more efficient and slightly more accurate to share the parameters of the feature encoder and the positional encoder between the teacher and student networks.

\paragraph{Targets.}
Training targets are constructed based on the output of the top $K$ blocks of the teacher network for time-steps which are masked in student-mode.\footnote{
We generally use the output of the FFN prior to the last residual connection in each block as target.
See the ablation in~\textsection\ref{sec:ablations}.
}
The output of block $l$ at time-step $t$ is denoted as $a_t^l$. 
We apply a normalization to each block to obtain $\hat{a}_t^l$ before averaging the top $K$ blocks $y_t = \frac{1}{K} \sum_{l=L-K+1}^L \hat{a}_t^l$ for a network with $L$ blocks in total to obtain the training target $y_t$ for time-step $t$.
This creates training targets that are to be regressed by the model when in student mode.
In preliminary experiments we found that averaging performed as well as predicting each block separately with a dedicated projection while enjoying the advantage of being more efficient.

Normalizing the targets helps prevent the model from collapsing into a constant representation for all time-steps and it also prevents layers with high norm to dominate the target features.
For speech representations, we use instance normalization~\citep{ulyanov2016in} without any learned parameters over the current input sample since neighboring representations are highly correlated due to the small stride over the input data, while for NLP and vision we found parameter-less layer normalization~\citep{ba2016layer} to be sufficient.
Variance-Invariance-Covariance regularization~\citep{bardes2021vicreg} also addresses this problem but we found the above strategy to perform well and it  does not introduce additional hyper-parameters.

\subsection{Objective}

Given contextualized training targets $y_t$, we use a Smooth L1 loss to regress these targets:
\begin{equation}
\mathcal{L}(y_t, f_t(x)) = 
\begin{cases}
\frac{1}{2} (y_t - f_t(x))^2/\beta & | y_t - f_t(x) | \leq \beta \\
(|y_t-f_t(x)|-\frac{1}{2}\beta) & \text{otherwise}
\end{cases}
\nonumber
\end{equation}
where $\beta$ controls the transition from a squared loss to an $L_1$ loss, depending on the size of the gap between the target $y_t$ and the model prediction $f_t(x)$ at time-step $t$. 
The advantage of this loss is that it is less sensitive to outliers, however, we need to tune the setting of $\beta$.

\section{Experimental Setup}
\label{sec:setup}

We experiment with two model sizes: \name{} Base and \name{} Large, containing either $L=12$ or $L=24$ Transformer blocks with $H=768$ or $H=1024$ hidden dimension (with $4\times H$ feed-forward inner-dimension). 
EMA updates are performed in fp32 for numerical stability~\citep{manohar2021kaizen}.

\subsection{Computer Vision}
\label{sec:setup_cv}

We embed images of 224x224 pixels as patches of 16x16 pixels~\citep{dosovitskiy2020vit}.
Each patch is linearly transformed and a sequence of 196 representations is input to a standard Transformer.
We follow BEiT~\citep{bao2021beit} by masking blocks of multiple adjacent patches where each block contains at least 16 patches with a random aspect ratio.
Different to their work, we found it more accurate to mask 60\% of the patches instead of 40\%.
We use randomly applied resized image crops, horizontal flipping, and color jittering~\citep{bao2021beit}. 
We use the same modified image both in teacher mode and student mode.

ViT-B models are pre-trained for 800 epochs.
As batch size we use 2,048 for ViT-B and 8,192 for ViT-L.
We use Adam~\citep{kingma2015adam} and a cosine schedule~\citep{loshchilov2016cosine} with a single cycle where we warm up the learning rate for 40 epochs to 0.002 for ViT-B and for 80 epochs to 0.001 for ViT-L after which the learning rate is annealed following the cosine schedule.
For ViT-B and ViT-L, we use $\beta=2$, $K=6$ and $\tau = 0.9998$ as a constant value with no schedule which worked well.
We use stochastic depth with rate 0.2~\citep{huang2016deep}.
For ViT-L, we train for 1,600 epochs in total, the first 800 epochs use $\tau=0.9998$, we then reset the learning rate schedule and the teacher weights to the student and continue for another 800 epochs with $\tau=0.9999$.

For image classification we mean-pool the output of the last Transformer block and input it to a softmax-normalized classifier.
We fine-tune ViT-B for 100 epochs and ViT-L for 50 epochs using Adam and a cosine schedule where we warmup up the learning rate for 20 epochs to 0.004 for ViT-B and for 5 epochs to 0.004 for ViT-L after which the learning rate follows the cosine schedule. 
We build on the open source implementation of BEiT~\citep{bao2021beit}.

\subsection{Speech Processing}
\label{sec:setup_speech}

Models are implemented in fairseq~\cite{ott2019fairseq} and take as input 16 kHz waveform which is processed by a feature encoder~\citep{baevski2020wav} containing seven temporal convolutions with 512 channels, strides (5,2,2,2,2,2,2) and kernel widths (10,3,3,3,3,2,2).
This results in an encoder output frequency of 50 Hz with a stride of about 20ms between each sample, and a receptive field of 400 input samples or 25ms of audio. 
The raw waveform input to the encoder is normalized to zero mean and unit variance.

The masking strategy for the Base model is also identical to~\citet{baevski2020wav}: we sample $p=0.065$ of all time-steps to be starting indices and mask the subsequent ten time-steps.
This results in approximately 49\% of all time-steps to be masked for a typical training sequence.
During pre-training we linearly anneal $\tau$ using $\tau_0=0.999$, $\tau_e=0.9999$ and $\tau_n=30,000$, average the top $K=8$ blocks as targets and found a simple L2 loss to work well.

We optimize with Adam~\citep{kingma2015adam}, with a peak learning rate of $\Enot{5e-4}$ for \name{} Base. 
The Base model uses a tri-stage scheduler which linearly warms up the learning rate over the first 3\% of updates, holds it for 90\% of updates and then linearly decays it over the remaining 7\%. 
We train \name{} Base for 400K updates with a batch size of 63 minutes of audio (61M frames).
We follow the fine-tuning regime of wav2vec 2.0~\citep{baevski2020wav} whose hyper-parameters depend on the labeled data setup.

\subsection{Natural Language Processing}
\label{sec:setup_nlp}

We build on the BERT re-implementation RoBERTa~\citep{liu2019roberta} available in fairseq~\citep{ott2019fairseq}.
The input data is tokenized using a byte-pair encoding~\citep{sennrich2016bpe} of 50K types and the model learns an embedding for each type~\citep{devlin2018bert,liu2019roberta} 
Once the data is embedded, we apply the BERT masking strategy to 15\% of uniformly selected tokens: 
80\% are replaced by a learned mask token, 10\% are left unchanged and 10\% are replaced by randomly selected vocabulary token; we do not use the next-sentence prediction task. 
We also consider the wav2vec 2.0 strategy of masking spans of four tokens.

For pre-training we use $\tau_0=0.999$, $\tau_e=0.9999$ and $\tau_n=100,000$, $K=10$ and set $\beta=4$.
The model is optimized with Adam over 1M updates using a tri-stage learning rate schedule (5\%, 80\% and 15\% of updates for warm-up, holding and linearly decaying, respectively). 
The peak learning rate is $\Enot{2e-4}$.
We train on 16 GPUs with a total batch size of 256 sequences and each sequence is up to 512 tokens.
For downstream tasks, we fine-tune the pre-trained model with four different learning rates ($\Enot{1e-5}$, $\Enot{2e-5}$, $\Enot{3e-5}$, $\Enot{4e-5}$) and choose the one which performs best across all considered NLP downstream tasks.

\section{Results}
\label{sec:results}

\subsection{Computer Vision}
\label{sec:results_cv}

\insertINtable

To evaluate our approach for computer vision, we pre-train \name{} on the images of the ImageNet-1K training set~\citep{deng2009imagenet} and fine-tune the resulting model for image classification using the labeled data of the same benchmark (\textsection\ref{sec:setup_cv}).
Following standard practice, models are evaluated in terms of top-1 accuracy on the validation set.
We distinguish between results based on a single self-supervised model, and results which train a separate visual tokenizer on additional data~\citep{bao2021beit} or distill other self-supervised models~\citep{dong2022peco}.

\insertLLtable

\autoref{tab:in} shows that \name{} outperforms prior work with ViT-B and ViT-L in the single model setting and all prior work for ViT-L.
Predicting contextualized latent representations in a masked prediction setup can perform very well compared to approaches which predict local targets such as  the original input pixels~\citep{he2021mae,xie2021simmim}, engineered image features~\citep{wei2021masked} or visual tokens~\citep{bao2021beit}.
It also outperforms prior self-distillation methods~\citep{caron2021dino} which regressed the final layer of the student network while inputting two different augmented versions of an image to the student and teacher networks.

\subsection{Speech and Audio Processing}
\label{sec:results_speech}

For speech processing, we pre-train \name{} on the 960 hours of speech audio data from \libri{} (\librisz{}). 
This dataset contains relatively clean speech audio from read audiobooks in English and is a standard benchmark in the speech community.
To get a sense of performance in different resource settings, we fine-tune models for automatic speech recognition using different amounts of labeled data, ranging from just 10 minutes to 960 hours.
We also compare to other work from the literature, including \wvpp{}~\citep{baevski2020wav} and HuBERT~\citep{hsu2020hubert}, two popular algorithms for speech representation learning relying on discrete units of speech.

\autoref{tab:librilight} shows improvements for most labeled data setups with the largest gains for 10 minutes of labeled data (20\% relative WER improvement) for the Base models.
For Large models, there are strong improvments for the smallest labeled data setups, and comparable performance for the resource-rich settings of 100 hours and 960 hours of labeled data where performance is generally saturating for many models~\citep{zhang2020pushing,chung2021w2vbert}.
Our results suggest that learning discrete units is not required when rich contextualized targets are used and that learning contextualized targets during pre-training improves performance.

\insertAStable

To further validate our approach for speech, we also trained a model on the AudioSet benchmark~\citep{gemmeke2017audioset}. 
The pre-training setup is the same as Librispeech, but we set $K=12$ and train for 200K updates with a batch size of 94.5 minutes of audio. 
We apply DeepNorm~\citep{wang2022deepnorm} and layer normalization of the targets to stabilize training. 
Finetuning is done on the balanced subset over 13k updates with a batch size of 21.3 minutes. 
Similar to~\citet{srivistava2021nonspeech}, we use linear softmax pooling~\citep{wang2018linsoftmax} and mixup~\citep{tokozume2017mixup} with probability 0.7. 
We add a single linear projection layer into 527 audioset classes and set the projection learning rate to $2e-4$. 
The pre-trained parameters are trained with a learning rate of $3e-5$ and we use masking during fine-tuning similar to~\citet{baevski2020wav} with $p=0.45$ and mask length of 4. 
The results (\autoref{tab:audioset}) show that data2vec can outperform a comparable setup that uses the same pre-training and fine-tuning data.

\subsection{Natural Language Processing}
\label{sec:results_nlp}

\insertGLUEtable

To get a sense of how \name{} performs for language, we adopt the same training setup as BERT~\citep{devlin2018bert} by pre-training on the Books Corpus~\citep{zhu2015books} and English Wikipedia data over 1M updates and a batch size of 256 sequences.
We evaluate on the General Language Understanding Evaluation (GLUE) benchmark~\citep{wang2018glue} which includes tasks for natural language inference (MNLI, QNLI, RTE), sentence similarity (MRPC, QQP and STS-B), grammaticality (CoLA), and sentiment analysis (SST-2).\footnote{MNLI (Multi Genre Natural Language Inference;~\citealt{williams2017mnli}), Stanford Question Answering Dataset (QNLI;~\citealt{rajpurkar2016squad}), Recognizing Textual Entailment~(RTE;\citealt{dagan2006rte1,haim2006rte2,giampiccolo2007rte3,bentivogli2009rte}), and we exclude Winograd NLI task from our results similar to~\citet{devlin2018bert},
Microsoft Research Paragraph Corpus (MRPC; \citealt{dolan2005mrpc}), Quora Question Pairs benchmark (QQP), and the Semantic Textual Similarity Benchmark (STS-B; \citealt{cer2017stsb}),
Corpus of Linguistic Acceptability (CoLA;~\citealt{warstadt2018cola}),
Stanford Sentiment Treebank (SST-2;~\citealt{socher2013sst2})}
We fine-tune \name{} separately on the labeled data provided by each task and report the average accuracy on the development sets over five fine-tuning runs.
We compare to the published BERT results as well as to the results we obtain by retraining RoBERTa in the current setup~(Baseline;~\citealt{liu2019roberta}) which provides a more suitable baseline to \name{} since we build on their open source code.

The results (\autoref{tab:glue}) show that \name{} outperforms the RoBERTa baseline.
When we mask spans of four BPE tokens with masking probability 0.35~\citep{baevski2020wav}, then results improve further.\footnote{We used a cosine learning rate schedule for this result.} 
This strategy does not leave tokens unmasked or uses random targets as for BERT (\textsection\ref{sec:setup_nlp}).

\insertLayerAblation

To our knowledge this is the first successful pre-trained NLP model which does not use discrete units (words, sub-words, characters or bytes) as the training target. 
Instead, the model predicts a \emph{contextualized latent representation} emerging from self-attention over the entire unmasked text sequence.
This enables a learning task where the model needs to predict targets with specific properties of the current text sequence rather than representations which are generic to every text sequence in which the particular discrete unit occurs.
Moreover, the set of training targets is not fixed, i.e., not a closed vocabulary, and the model can choose to define new targets as it sees fit, akin to an open vocabulary setting.

\insertContextAblation

\subsection{Ablations}
\label{sec:ablations}

\paragraph{Layer-averaged Targets.}
One of the main differences of our method compared to BYOL is the use of targets which are based on averaging multiple layers from the teacher network (\textsection\ref{sec:method_targets}).
This idea was partly inspired by the fact that the top layers of wav2vec 2.0 do not perform as well for downstream tasks as layers in the middle of the network~\citep{baevski2021unsupervised,pasad2021layerwise}.

In the next experiment, we measure performance for all three modalities when averaging $K=1,\dots,12$ layers where $K=1$ corresponds to predicting only the top layer similar to BYOL.
For faster experimental turn-around, we train Base models with $L=12$ layers in total.
For speech, we pre-train for 200K updates on \libri{}, fine-tune on the 10 hour labeled split of Libri-light~\citep{librilight} and report word error rate without a language model on dev-other.
For NLP, we report the average GLUE score on the validation set (\textsection\ref{sec:results_nlp}) and for computer vision we pre-train models for 300 epochs and report the top-1 accuracy on ImageNet (\textsection\ref{sec:results_cv}).

\autoref{fig:layer_ablation} shows that targets based on multiple layers improves over using only the top layer ($K=1$) for all modalities.
Using all layers is generally a good choice and only slightly worse than a carefully tuned value of $K$.
Neural networks build features over multiple layers and different types of features are extracted at different layers.
Using features from multiple layers enriches the self-supervised task and improves accuracy.

\paragraph{Target Contextualization.}
Teacher representations are based on self-attention over the entire input which results in contextualized targets.
This distinguishes data2vec from other self-supervised approaches which construct a learning task by predicting or reconstructing local parts of the input (\textsection\ref{sec:related}).
This poses the natural question of whether contextualized targets are required for data2vec to work well.

In order to get a better sense of this, we construct target representations which do not have access to the entire input sample but rather only a pre-defined fraction of it.
Concretely, we restrict the self-attention mechansim of the teacher to only be able to access a portion of the input surrounding the current time-step.
Once the model is trained, we fine-tune it so that it can access the full context size.
\autoref{fig:context_ablation} shows that larger context sizes lead to better downstream performance. 
The best accuracy is achieved when the entire input sample is visible.
This shows that richer target representations can indeed lead to better performance.

\insertTargetFeatureAblation

\paragraph{Target Feature Type.}
Transformers blocks contain several layers which can each serve as targets.
To get a sense of how different layers impact performance, we pre-train speech models on \libri{} using different layers as target features.
\autoref{tab:target_location} shows that the output of the feedforward network (FFN) block works best while the output of the self-attention block does not yield a usable model.
We believe this is because the self-attention is before the residual connection and features are heavily biased towards other time-steps. 
This issue is alleviated by the use of FFN features since these also include the features before the self-attention.

\section{Discussion}

\paragraph{Modality-specific Feature Extractors and Masking.}
Our primary goal is to design a single learning mechanism for different modalities. 
Despite the unified learning regime, we still use modality-specific feature extractors and masking strategies.
This makes sense given the vastly different nature of the input data:
for example, in speech we learn from a very high resolution input (16 kHz waveform) which contains hundreds of thousands of samples for typical utterances. 
To process this, we apply a multilayer convolutional neural network to obtain a 50 Hz feature sequence.
For NLP, word input sequences have vastly lower resolution which can be directly embedded in distributional space via a lookup table.
The type of data also impacts how we should mask the input to create a challenging learning task:
removing individual words provides a sufficiently challenging task but for speech it is necessary to mask spans since adjacent audio samples are highly correlated with each other.
Relatedly, there has been recent work on a Transformer architecture that can directly operate on the raw data of different modalities without modality-specific feature encoders~\citep{jaegle2021perceiver,jaegle2022perceiverio}.
Their work is focused on supervised learning for classification tasks and we believe that our work is complementary.

\paragraph{Structured and Contextualized Targets.}
One of the main differences of \name{} to most other masked prediction work~\citep{devlin2018bert,baevski2020wav,ling2020decoar,bao2021beit,he2021mae,wei2021masked} is that the features of the training targets are contextualized since the features are built with self-attention over the entire unmasked input in teacher mode.
And while BYOL~\citep{grill2020byol} and DINO~\citep{caron2021dino} also use latent target representations based on the entire input, their focus is on learning transformation-invariant representations instead of structural information within a sample.

One exception is HuBERT~\citep{hsu2020hubert} which builds a fixed set of discrete target units by clustering Transformer layer representations.
In comparison, \name{} has no limitation on the number of target units.
Instead of representing each instance of particular discrete target unit with the same set of features, \name{} can build target features that are specific to the current sequence.

For NLP, we believe~\name{} is the first work that does not rely on pre-defined target units.
Most other work uses either words, sub-words~\citep{radford2018unsup,devlin2018bert}, characters~\citep{tay2021charformer} or even bytes~\citep{xue2021byt5}.
Aside, defining word boundaries is not straightforward for some  Asian languages.
Contextualized targets enable integrating features from the entire sequence into the training target which provides a richer self-supervised task. 
Furthermore, the representation of each instance of a particular unit (word/sub-word/character/byte) can differ for the masked prediction task.
This enables to associate a different meaning to a particular depending on the context it occurs in.
It also relieves the model from the need to learn a single set of features for a target unit that fits all instances of this unit.

\paragraph{Representation Collapse.}
A common failure mode when learning targets is for similar representations for targets to be produced resulting in a trivial task~\citep{jing2021understanding}.
To deal with this, contrastive models such as wav2vec 2.0~\citep{baevski2020wav} use the same target representation both as a positive and a negative example.
BYOL~\citep{grill2020byol} do not optimize the teacher parameters to minimize the loss and VicReg~\citep{bardes2021vicreg} adds an explicit loss encouraging variance among different representations.

We found that collapse is most likely to happen in the following scenarios: 
First, the learning rate is too large or the learning rate warmup is too short which can often be solved by tuning the respective hyperparameters.
Second, $\tau$ is too low which leads to student model collapse and is then propagated to the teacher. 
This can be addressed by tuning $\tau_0$, $\tau_e$ and $\tau_n$.
Third, we found collapse to be more likely for modalities where adjacent targets are very correlated and where longer spans need to be masked, e.g., speech.
We address this by promoting variance through normalizing target representations over the sequence or batch~\citep{grill2020byol}.
For models where targets are less correlated, such as vision and NLP, momentum tracking is sufficient. 

\section{Conclusion}

Recent work showed that uniform model architectures can be effective for multiple modalities~\citep{jaegle2021perceiver}.
Whereas we show that a single self-supervised learning regime can be effective for vision, speech and language.
The key idea is to regress contextualized representations based on a partial input view.
\name{} outperforms prior self-supervised work on ImageNet-1K for ViT-B and ViT-L single models, it improves over prior work on speech recognition for the low-resource Libri-light setups, and it outperforms RoBERTa on GLUE in the original BERT setup.

A single learning method for multiple modalities will make it easier to learn across modalities and future work may investigate tasks such as audio-visual speech recognition or cross-modal retrieval.
Our approach still uses modality-specific input encoders and we adopt modality-specific masking strategies which future work may unify.

\section*{Acknowledgements}

We thank Brenden Lake, Dhruv Batra, Marco Baroni and Laurens van der Maaten for helpful discussions.

\bibliography{refs}
\bibliographystyle{icml2022}

\newpage
\appendix
\onecolumn
\section{Extended speech processing results}

\insertLLtableApp


\section{Comparison of loss functions}

\autoref{tab:loss_ablation} shows that different choices of the loss function have a relatively small effect on final performance.

\begin{table}[h]
\centering
\caption{Different pre-training losses on Librispeech dev-other (no language model). 
\label{tab:loss_ablation}}
\begin{tabular}[t]{lr}
\toprule
& WER \\
\midrule
L2 & 17.1 \\
L1 & 17.2 \\
Smooth L1 ($\beta=0.08$) & 17.2 \\
Smooth L1 ($\beta=0.25$) & 16.8 \\
Smooth L1 ($\beta=0.5$) & 16.8 \\
Smooth L1 ($\beta=1$) & 17.3 \\
\bottomrule
\end{tabular}
\end{table}

\section{Speech masking parameter ablation}

Our method requires different masking parameters for each modality and this makes intuitive sense: masking 15\% of inputs is effective for text but not for images since text tokens are highly semantic and it is sufficient to mask a smaller proportion of the input to construct a useful task.
We rely largely on settings from the literature, except for images, where we found that a higher masking rate compared to BEiT works slightly better.
When we tried tuning masking hyperparameters for other modalities than vision, we did not see significant improvements (e.g., see \autoref{tab:speech_masking_ablation}) for speech. 
The small improvements we saw are likely to disappear after adding a language model.

\begin{table}[h]
\centering
\caption{Ablation of speech masking parameters. Results are on Librispeech dev-other without a language model. 
\label{tab:speech_masking_ablation}}
\begin{tabular}[t]{lr}
\toprule
& WER \\
\midrule
baseline (mask prob = 0.65, mask len = 10) & 17.1 \\
mask prob = 0.8 & 17.2 \\
mask prob = 0.5 & 17.3 \\
mask len = 5 & 22.1 \\
mask len = 15 & 18.9 \\
\bottomrule
\end{tabular}
\end{table}

\end{document}

%% file: macro.tex
\newcommand{\name}{data2vec}

\newcommand{\wvpp}{wav2vec 2.0}
\newcommand{\wvppbase}{\wvpp{} Base}

\newcommand{\hubase}{Hubert Base}

\newcommand{\libri}{Librispeech}
\newcommand{\libril}{Libri-light}
\newcommand{\voxsz}{LL-60K}
\newcommand{\librisz}{LS-960}
\newcommand{\libriunsz}{LS-860}

\newcommand{\Enot}[1]{\num[exponent-product = \times]{#1}}

%% file: tables.tex
\newcommand{\insertINtable}{
\begin{table}[t]
\centering
\caption{Computer vision: top-1 validation accuracy on ImageNet-1K with ViT-B and ViT-L models. 
\name{} ViT-B was trained for 800 epochs and ViT-L for 1,600 epochs.
We distinguish between individual models and setups composed of multiple models (BEiT/PeCo train separate visual tokenizers and PeCo also distills two MoCo-v3 models).
\label{tab:in}}
\begin{tabular}[t]{lrrrrr}
\toprule
& ViT-B & ViT-L \\
\midrule
\textit{Multiple models} \\
BEiT~\citep{bao2021beit} & 83.2 & 85.2 \\
PeCo~\citep{dong2022peco} & 84.5 & 86.5 \\
\midrule
\textit{Single models} \\
MoCo v3~\citep{chen2021mocov3} & 83.2 & 84.1 \\
DINO~\citep{caron2021dino} & 82.8 & - \\
MAE~\citep{he2021mae} & 83.6 & 85.9 \\
SimMIM~\citep{xie2021simmim} & 83.8 & - \\
iBOT~\citep{zhou2021ibot} & 83.8 & - \\
MaskFeat~\citep{wei2021masked} & 84.0 & 85.7 \\
\name{} & 84.2 & 86.6 \\
\bottomrule
\end{tabular}
\end{table}
}

\newcommand{\insertAStable}{
\begin{table}[h!]
\caption{Audio event classification: mean average precision (mAP) on the eval set of AudioSet when pre-training on all of AudioSet (AS) and/or Librispeech (LS) and fine-tuning on the 20K subset.
\label{tab:audioset}
}
\vspace{0.075in}
\centering 
\begin{tabular}{lrr}
\toprule
 & Pretrain data & mAP \\
\midrule
SSAST~\citep{gong2021ssast} & AS, LS & 31.0 \\
MaskSpec~\citep{chong2022maskspec} & AS & 32.3 \\
MAE-AST~\citep{baade2022} & AS, LS & 30.6 \\
\midrule
data2vec & AS & 34.5 \\
\bottomrule
\end{tabular}
\end{table}
}

\newcommand{\insertLLtable}{
\begin{table*}[h!]
\caption{Speech processing: word error rate on the \libri{} test-other when fine-tuning pre-trained models on the \libril{} low-resource labeled data setups~\citep{kahn2020librilight} of 10 min, 1 hour, 10 hours, the clean 100h subset of \libri{} and the full 960h of \libri{}. 
For pretraining, models use 960 hours of unlabeled audio from \libri{} (\librisz{}), or the 60K hours from \libril{} (\voxsz{}); WavLM Large uses 94K hours (MIX-94K) which includes \voxsz{} as well as other datasets.
We indicate the language model used during decoding (LM).
Results for all dev/test sets and other LMs can be found in the supplementary material (\autoref{tab:librilight_app}).
\label{tab:librilight}
}
\vspace{0.075in}
\centering 
\begin{tabular}{lccrrrrr}
\toprule
& Unlabeled & \multirow{2}{*}{LM} & \multicolumn{5}{c}{Amount of labeled data} \\
{} & data & {} & 10m & 1h & 10h & 100h & 960h \\
\midrule
\textit{Base models} \\
\wvpp{}~\citep{baevski2020wav} & \librisz{} & 4-gram & 15.6 & 11.3 & 9.5 & 8.0 & 6.1 \\
HuBERT~\citep{hsu2020hubert} & \librisz{} & 4-gram & 15.3 & 11.3 & 9.4 & 8.1 & - \\
WavLM~\citep{chen2021wavlm} & \librisz{} & 4-gram & - & 10.8 & 9.2 & 7.7 & - \\
\midrule
\name{} & \librisz{} & 4-gram & 12.3 & 9.1 & 8.1 & 6.8 & 5.5 \\
\midrule
\textit{Large models} \\
\wvpp{}~\citep{baevski2020wav} & \voxsz{} & 4-gram & 10.3 & 7.1 & 5.8 & 4.6 & 3.6 \\
HuBERT~\citep{hsu2020hubert} & \voxsz{} & 4-gram & 10.1 & 6.8 & 5.5 & 4.5 & 3.7 \\
WavLM~\citep{chen2021wavlm} & MIX-94K & 4-gram & - & 6.6 & 5.5 & 4.6 & - \\
\midrule
\name{} & \voxsz{} & 4-gram & 8.4 & 6.3 & 5.3 & 4.6 & 3.7 \\
\bottomrule
\end{tabular}
\end{table*}
}

\newcommand{\insertLLtableApp}{
\begin{table*}[h!t]
\caption{Speech processing: word error rate on the \libri{} dev/test sets when training on the \libril{} low-resource labeled data setups of 10 min, 1 hour, 10 hours and the clean 100h subset of \libri{}. 
Models use the audio of \libri{} (\librisz{}) as unlabeled data.
\label{tab:librilight_app}
}
\centering 
\begin{tabular}{lcccrrrr}
\toprule
\multirow{2}{*}{Model} & Unlabeled & \multirow{2}{*}{LM} & \multicolumn{2}{c}{dev} && \multicolumn{2}{c}{test} \\
\cline{4-5}\cline{7-8} 
{} & data & {} & clean & other && clean & other \\
\midrule
\midrule
\multicolumn{8}{l}{\textbf{10 min labeled}}\\
\wvppbase{}~\citep{baevski2020wav} & \librisz{} & 4-gram & 8.9 & 15.7 && 9.1 & 15.6 \\
\hubase{}~\citep{hsu2020hubert} & \librisz{} & 4-gram & 9.1 & 15.0 && 9.7 & 15.3 \\ 
\midrule
\name{} Base & \librisz{} & 4-gram & 7.3 & 11.6 && 7.9	& 12.3 \\
\midrule
\midrule
\multicolumn{8}{l}{\textbf{1h labeled}}\\
\wvppbase{}~\citep{baevski2020wav} & \librisz{} & 4-gram & 5.0 & 10.8 && 5.5 & 11.3 \\
\hubase{}~\citep{hsu2020hubert} & \librisz{} & 4-gram & 5.6 & 10.9 && 6.1 & 11.3 \\
\midrule
\name{} Base & \librisz{} & 4-gram & 4.0 & 8.5 && 4.6 & 9.1 \\
\midrule
\midrule
\multicolumn{8}{l}{\textbf{10h labeled}}\\
\wvppbase{}~\citep{baevski2020wav} & \librisz{} & 4-gram & 3.8 & 9.1 && 4.3 & 9.5 \\
\hubase{}~\citep{hsu2020hubert} & \librisz{} & 4-gram & 3.9 & 9.0 && 4.3 & 9.4 \\
\midrule
\name{} Base & \librisz{} & 4-gram & 3.3 & 7.5 && 3.9 & 8.1 \\
\midrule
\midrule
\multicolumn{8}{l}{\textbf{100h labeled}}\\
Noisy student~\citep{park2020improved} & \libriunsz{} & LSTM & 3.9 & 8.8 && 4.2 & 8.6 \\
IPL~\citep{xu2020iterative} & \voxsz{} & 4-gram+Transf. & 3.2 & 6.1 && 3.7 & 7.1 \\
SlimIPL~\citep{likhomanenko2021slimipl} & \libriunsz{} & 4-gram+Transf. & 2.2 & 4.6 && 2.7 & 5.2 \\
\wvppbase{}~\citep{baevski2020wav} & \librisz{} & 4-gram & 2.7 & 7.9 && 3.4 & 8.0 \\
\hubase{}~\citep{hsu2020hubert} & \librisz{} & 4-gram & 2.7 & 7.8 && 3.4 & 8.1 \\
\midrule
\name{} Base & \librisz{} & 4-gram & 2.2 & 6.4 && 2.8 & 6.8 \\
\bottomrule
\end{tabular}
\end{table*}
}

\newcommand{\insertGLUEtable}{
\begin{table*}[h!]
\centering
\caption{Natural language processing: GLUE results on the development set for single-task fine-tuning of individual models.
For MNLI we report accuracy on both the matched and unmatched dev sets, for MRPC and QQP, we report the unweighted average of accuracy and F1, for STS-B the unweighted average of Pearson and Spearman correlation, for CoLA we report Matthews correlation and for all other tasks we report accuracy.
BERT Base results are from~\citet{wu2020clear} and our baseline is RoBERTa re-trained in a similar setup as BERT.
We also report results with wav2vec 2.0 style masking of spans of four BPE tokens with no unmasked tokens or random targets.
\label{tab:glue}} 
\vspace{0.075in}
\begin{tabular}{lrrrrrrrrr}
\toprule
& MNLI & QNLI & RTE & MRPC & QQP & STS-B & CoLA & SST & Avg.\\
\midrule
BERT~\citep{devlin2018bert} & 84.0/84.4 & 89.0 & 61.0 & 86.3 & 89.1 & 89.5 & 57.3 & 93.0 & 80.7\\
Baseline~\citep{liu2019roberta} & 84.1/83.9 & 90.4 & 69.3 & 89.0 & 89.3 & 88.9 & 56.8 & 92.3 & 82.5 \\ 
\midrule
\name{} & 83.2/83.0 & 90.9 & 67.0 & 90.2 & 89.1 & 87.2 & 62.2 & 91.8 & 82.7\\
~+ wav2vec 2.0 masking & 82.8/83.4 & 91.1 & 69.9 & 90.0 & 89.0 & 87.7 & 60.3 & 92.4 & 82.9 \\
\bottomrule
\end{tabular}
\end{table*}
}

\pgfplotstableread[row sep=\\,col sep=&]{
lyr & vitb & speechb & nlpb \\
1 & 83.01 & 39.39 & 58.63 \\
2 & 83.5 & 36.04 & 73.08 \\
3 & 84.02 & 26.37 & 76.80 \\
4 & 83.88 & 26.55 & 78.03 \\
5 & 83.86 & 17.43 & 79.09 \\
6 & 83.79 & 13.73 & 80.07 \\
7 & 83.74 & 13.90 & 80.15 \\
8 & 83.7 & 13.43 & 81.42 \\
9 & 83.64 & 15.44 & 80.50 \\
10 & 83.74 & 13.74 & 82.13 \\
11 & 83.65 & 14.01 & 81.44 \\
12 & 83.66 & 14.08 & 81.63 \\
}\layerAblationData

\newcommand{\insertLayerAblation}{
\begin{figure*}
\centering
\begin{subfigure}[b]{.33\textwidth}
    \centering
    \begin{tikzpicture}
    \begin{axis}[
    width=1\textwidth,
    height=.75\textwidth,
    legend style={font=\small,
    at={(0.98,0.27)},
    anchor=east,legend columns=1},
    xticklabels from table={\layerAblationData}{lyr},
    xticklabel style={font=\scriptsize},
    xtick=data,
    yticklabel style={font=\small},
    ylabel={Word error rate},
    ylabel style={font=\small},
    ylabel near ticks, 
    xlabel={$K$},
    xlabel style={font=\small},
    ]
    \addplot[blue,mark=*,mark options={solid,scale=1,fill=blue}] table[x expr=\coordindex,y=speechb]{\layerAblationData};
    \end{axis}
    \end{tikzpicture}
    \caption{Speech}
    \label{fig:layer_ablation_speech}
\end{subfigure}
\begin{subfigure}[b]{.33\textwidth}
    \centering
    \begin{tikzpicture}
    \begin{axis}[
    width=1\textwidth,
    height=.75\textwidth,
    legend style={font=\small,
    at={(0.98,0.27)},
    anchor=east,legend columns=1},
    xticklabels from table={\layerAblationData}{lyr},
    xticklabel style={font=\scriptsize},
    xtick=data,
    yticklabel style={font=\small},
    ylabel={GLUE score},
    ylabel style={font=\small},
    ylabel near ticks, 
    xlabel={$K$},
    xlabel style={font=\small},
    ]
    \addplot[red,mark=*,mark options={solid,scale=1,fill=red}] table[x expr=\coordindex,y=nlpb]{\layerAblationData};
    \end{axis}
    \end{tikzpicture}
    \caption{NLP}
    \label{fig:layer_ablation_nlp}
\end{subfigure}
\begin{subfigure}[b]{.33\textwidth}
    \centering
    \begin{tikzpicture}
    \begin{axis}[
    width=1\textwidth,
    height=.75\textwidth,
    legend style={font=\small,
    at={(0.98,0.27)},
    anchor=east,legend columns=1},
    xticklabels from table={\layerAblationData}{lyr},
    xticklabel style={font=\scriptsize},
    xtick={0,1,2,3,4,5,6,7,8,9,10,11,12},
    yticklabel style={font=\small},
    xlabel style={font=\small},
    ymin=80,ymax=85,
    ylabel={Top-1 valid accuracy},
    ylabel near ticks, 
    xlabel={$K$},
    ylabel style={font=\small},
    ]
    \addplot[black,dashed,mark=*,mark options={solid,scale=1,fill=black}] table[x expr=\coordindex,y=vitb]{\layerAblationData};
    \end{axis}
    \end{tikzpicture}
    \caption{Vision}
    \label{fig:layer_ablation_vision}
\end{subfigure}
\caption{Predicting targets which are the average of multiple layers is more robust than predicting only the top most layer ($K=1$) for most modalities. 
We show the performance of predicting the average of $K$ teacher layer representations (\textsection\ref{sec:method_targets}).
The effect is very pronounced for speech and NLP while for vision there is still a slight advantage of predicting more than a single layer.
}
\label{fig:layer_ablation}
\end{figure*}
}

\newcommand{\insertTargetFeatureAblation}{
\begin{table}[t]
\centering
\caption{Effect of using different features from the teacher model as targets: 
we compare using the output of the self-attention module, the feed-forward module (FFN) as well as after the final residual connection (FFN + residual) and layer normalization (End of block).
Results are not directly comparable to the main results since we use a reduced setup (\textsection\ref{sec:ablations}).
}
\label{tab:target_location}
\begin{tabular}[t]{lrrrrr}
\toprule
Layer & WER \\
\midrule
self-attention & 100.0 \\
FFN & 13.1 \\
FFN + residual & 14.8 \\
End of block & 14.5 \\
\bottomrule
\end{tabular}
\end{table}
}

\pgfplotstableread[row sep=\\,col sep=&]{
ctxt & sec & perc & speechb \\
2 & 0.04 & 0.3 & 78.608 \\
5 & 0.1 & 0.8 & 63.255 \\
10 & 0.2 & 1.6 & 29.777 \\
25 & 0.5 & 4.1 & 23.526 \\
50 & 1 & 8.1 & 19.681 \\
100 & 2 & 16.2 & 15.859 \\
all & all & 100 & 14.807 \\
}\layerAblationContextSpeech

\pgfplotstableread[row sep=\\,col sep=&]{
ctxt & pixels & perc & vitb \\
2 & 512 & 1 & 76.96 \\
5 & 1280 & 2.6 & 79.34 \\
10 & 2560 & 5.1 & 80.79 \\
25 & 6400 & 12.8 & 82.29 \\
50 & 12800 & 25.5 & 83.1 \\
100 & 25600 & 51 & 83.57 \\
all & 50176 & 100 & 83.79 \\
}\layerAblationContextVision

\newcommand{\insertContextAblation}{
\begin{figure}
\centering
\begin{subfigure}[b]{.35\textwidth}
    \centering
    \begin{tikzpicture}
    \begin{axis}[
    width=1\textwidth,
    height=.75\textwidth,
    legend style={font=\small,
    at={(0.98,0.27)},
    anchor=east,legend columns=1},
    xticklabels from table={\layerAblationContextSpeech}{perc},
    xticklabel style={font=\scriptsize},
    xtick=data,
    yticklabel style={font=\small},
    ylabel style={font=\small},
    ylabel={Word error rate},
    ylabel near ticks, 
    xlabel style={font=\small},
    xlabel={Teacher context size (\%)},
    ]
    \addplot[blue,mark=*,mark options={solid,scale=1,fill=blue}] table[x expr=\coordindex,y=speechb]{\layerAblationContextSpeech};
    \end{axis}
    \end{tikzpicture}
    \caption{Speech}
    \label{fig:context_ablation_speech}
\end{subfigure}
\begin{subfigure}[b]{.35\textwidth}
    \centering
    \begin{tikzpicture}
    \begin{axis}[
    width=1\textwidth,
    height=.75\textwidth,
    legend style={font=\small,
    at={(0.98,0.27)},
    anchor=east,legend columns=1},
    xticklabels from table={\layerAblationContextVision}{perc},
    xticklabel style={font=\scriptsize},
    xtick={0,1,2,3,4,5,6,7,8,9,10,11,12},
    yticklabel style={font=\small},
    ylabel style={font=\small},
    ylabel={Top-1 valid accuracy},
    ylabel near ticks, 
    xlabel style={font=\small},
    xlabel={Teacher context size (\%)},
    ]
    \addplot[black,dashed,mark=*,mark options={solid,scale=1,fill=black}] table[x expr=\coordindex,y=vitb]{\layerAblationContextVision};
    \end{axis}
    \end{tikzpicture}
    \caption{Vision}
    \label{fig:context_ablation_vision}
\end{subfigure}
\caption{
Contextualized target representations improve accuracy. 
We mask all but a fraction of the data when constructing the target representations in the teacher for pre-training and report downstream performance for speech and vision.
Time-steps for speech are more fine-grained (20ms patches) compared to vision (16x16 pixels per patch).
}
\label{fig:context_ablation}
\end{figure}
}